\documentclass[10pt,twocolumn,letterpaper]{article}

\usepackage[pagenumbers]{cvpr} 

\usepackage{amsmath}
\usepackage{array}
\usepackage{tikz}
\usepackage{multirow}
\usepackage{caption}
\usepackage{placeins}
\usepackage{stfloats}

\usepackage{graphicx}

\definecolor{orange}{rgb}{1,0.5,0}
\definecolor{green}{rgb}{0,0.5,0}

\newif\ifcameraready 

\ifcameraready

\fi

\usepackage{soul}
\setuldepth{foobar}

\usepackage[ruled,linesnumbered]{algorithm2e}

\definecolor{cvprblue}{rgb}{0.21,0.49,0.74}
\usepackage[pagebackref,breaklinks,colorlinks,allcolors=cvprblue]{hyperref}

\def\confName{CVPR}
\def\confYear{2026}

\title{DOGE: Differentiable B\'{e}zier Graph Optimization for Road Network Extraction}

\author{Jiahui Sun \quad Junran Lu \quad Jinhui Yin \quad Yishuo Xu \quad Yuanqi Li \quad Yanwen Guo\thanks{Corresponding author.}\\
Nanjing University, Nanjing, China\\
{\tt\small \{cgjiahui, junranlu, 522025330134, yishuoxu\}@smail.nju.edu.cn} \quad {\tt\small \{yuanqili, ywguo\}@nju.edu.cn}
}

\begin{document}
\maketitle
\begin{abstract}
Automatic extraction of road networks from aerial imagery is a fundamental task, yet prevailing methods rely on polylines that struggle to model curvilinear geometry.
We maintain that road geometry is inherently curve-based and introduce the B\'{e}zier Graph, a differentiable parametric curve-based representation. The primary obstacle to this representation is to obtain the difficult-to-construct vector ground-truth (GT). 
We sidestep this bottleneck by reframing the task as a global optimization problem over the B\'{e}zier Graph. Our framework, \textit{DOGE}, operationalizes this paradigm by learning a parametric B\'{e}zier Graph directly from segmentation masks, eliminating the need for curve GT. 
\textit{DOGE} holistically optimizes the graph by alternating between two complementary modules: \textit{DiffAlign} continuously optimizes geometry via differentiable rendering, while \textit{TopoAdapt} uses discrete operators to refine its topology.
Our method sets a new state-of-the-art on the large-scale SpaceNet and CityScale benchmarks, presenting a new paradigm for generating high-fidelity vector maps of road networks.
We will release our code and related data.
\end{abstract}



\section{Introduction}
\label{sec:introduction}

Road network graphs, a form of standard-definition (SD) map, represent the geometry and topology of drivable roads. 
They are foundational to critical applications like route planning, autonomous driving, and urban modeling, where up-to-date and accurate data is paramount~\cite{Etten2018SpaceNetAR, demir2018DeepGlobe, tang2019CityFlow, chang2019Argoverse, sun2020Waymo, wang2017TorontoCity, xie2025citydreamer4d}. 
However, manually creating these maps remains a costly and labor-intensive bottleneck~\cite{mnih2010LearningRoadsECCV, maggiori2017InriaAerial, zamir2019iSAID}. The growing abundance of high-resolution satellite imagery presents a powerful alternative: automatically reconstructing road networks to enable rapid, city-scale updates at a low cost.


Most existing methods rely on a polyline representation, approximating road centerlines with sequences of connected line segments, as shown in \cref{fig:polyline-vs-curve}. This representation has fundamental limitations: polylines are inefficient, requiring a high density of vertices to accurately model curves; they are cumbersome to edit, as local changes can disrupt geometric smoothness; and only guaranteeing positional (C0) continuity~\cite{foley1996computer, chen2022road, bastaniRoadTracerAutomaticExtraction2018, hetangSegmentAnythingModel2024, zhangFastParallelAlgorithm1984b}.
\definecolor{customRed}{HTML}{e25241}
\definecolor{customYellow}{HTML}{fff455}
\definecolor{customBlue}{HTML}{3b82f6}   
\begin{figure}[t]
    \centering
    \begin{tikzpicture}
        \node[anchor=south west,inner sep=0] (image) at (0,0) {\includegraphics[width=\linewidth]{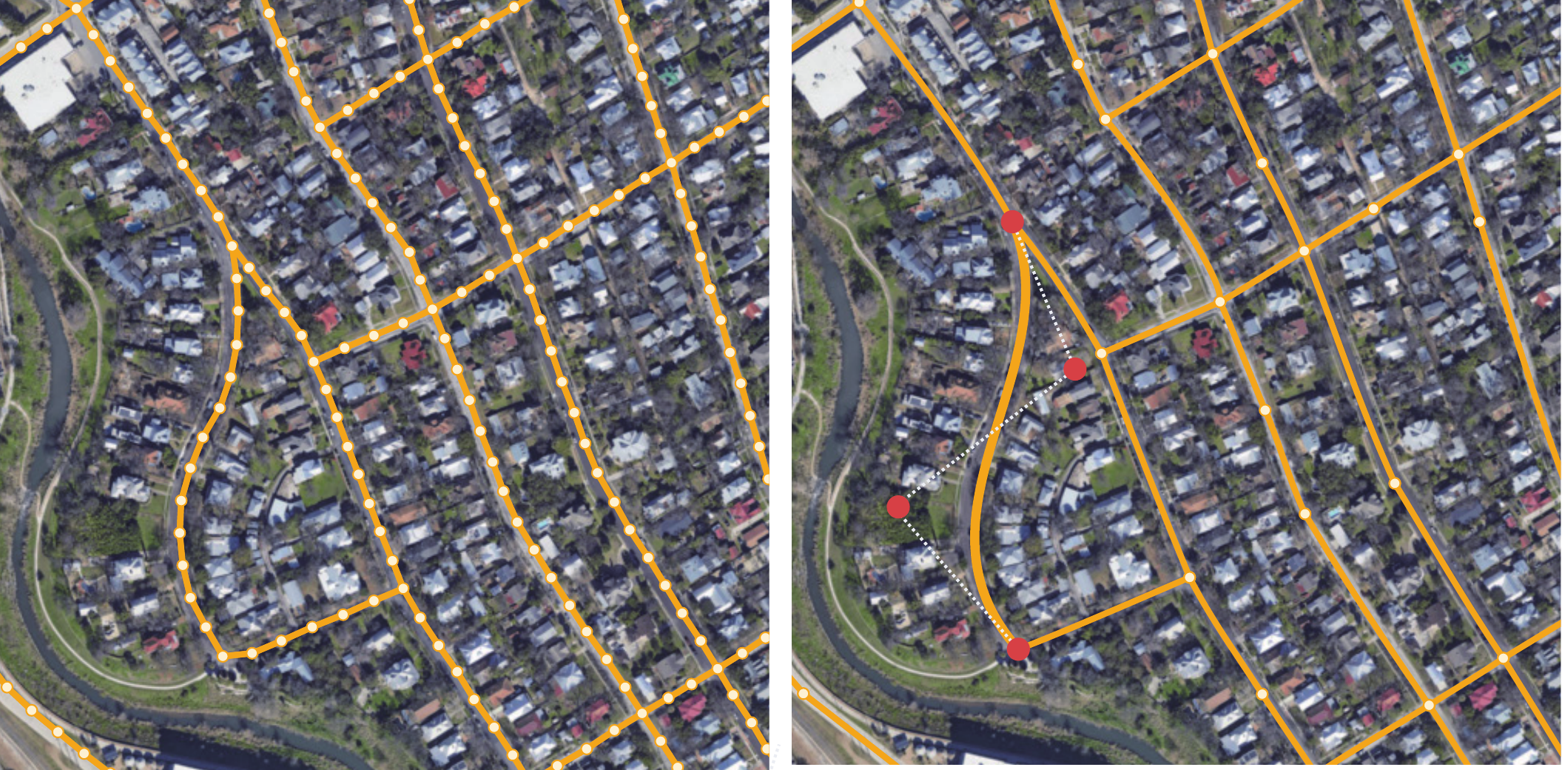}};
        
        \begin{scope}[x={(image.south east)},y={(image.north west)}]
            \node[anchor=north west] at (0.03, 0.01) {\footnotesize(a) Polyline Approximation};
            \node[anchor=north west] at (0.52, 0.01) {\footnotesize(b) B\'{e}zier Graph Representation};
        \end{scope}
    \end{tikzpicture}
    \vspace{-5mm} 
    \caption{Polyline versus curve-based road representations. (a) A polyline approximates a curve with discrete polylines. (b) A parametric B\'{e}zier curve representation. A road segment is shown 
    with its four control points (\textcolor{customRed}
    {red}) that define its geometry.}
    \label{fig:polyline-vs-curve}
    \vspace{-1mm}
\end{figure}



Our motivation is simple: road geometry is inherently curvilinear (\cref{fig:polyline-vs-curve}). Thus, we introduce the B\'{e}zier Graph, to our knowledge the first parametric representation for SD road networks using cubic B\'{e}zier curves. This representation is inherently smooth and analytically differentiable, enabling intuitive geometric editing via a few control points while preserving topological structure. It can be uniformly sampled to polylines at arbitrary resolution to remain compatible with existing pipelines. As shown in \cref{subsec:compactness_analysis}, our curve-based representation better models road networks with fewer nodes and edges than polyline methods, which substantially reduces the computational complexity. 

A conventional strategy to reconstruct the B\'{e}zier Graph would be to train a model to predict it directly. However, such an approach is immediately confronted by a significant bottleneck: the difficulty of creating the necessary vector GT. 
This challenge is twofold. First, the vectorization process is inherently ambiguous, leading to many equally valid curve factorizations for any given road network~\cite{bessmeltsev2019vectorization,reddy2021im2vec}. And relying on a rule-based heuristic transformation algorithm could encode bias in the label. 
Second, the conversion process can often be complex and brittle. Existing pipelines~\cite{blayneyBezierEverywhereAll2024} rely on multi-stage, heuristic-driven procedures including node selection, path splitting, and least‑squares fitting with global consistency.
The thresholds during the conversion are dataset‑dependent and difficult to standardize at scale.

We sidestep these challenges with \textit{DOGE} (\textbf{D}ifferentiable \textbf{O}ptimization of a B\'{e}zier \textbf{G}raph for road network \textbf{E}xtraction), a framework that reconstructs road networks by optimizing a B\'{e}zier Graph directly against segmentation masks, thus eliminating the need for vector GT.
Instead of building the graph with fixed, sequential decisions, \textit{DOGE} treats the road network as a parametric graph composed of differentiable curves (a B\'{e}zier graph) and holistically refines it against a segmentation mask. This reframes road extraction as a global optimization problem over the graph's continuous geometry and discrete topology.
\textit{DOGE} accomplishes this by decoupling the graph's geometry and topology. It addresses them with two complementary modules: \textit{DiffAlign} pioneers the use of differentiable rendering for continuous geometric alignment, while \textit{TopoAdapt} applies discrete operators to evolve the graph's topology. By alternating between them, \textit{DOGE} achieves a global, iterative refinement of the graph's shape and connectivity.

In practice, we use a fine-tuned SAM2~\cite{raviSAM2Segment2024, kirillovSegmentAnything2023} for high-quality segmentation supervision. Without requiring vector GT, our method learns a curve-based representation and sets a new state-of-the-art on the large-scale SpaceNet and CityScale benchmarks. We hope our work will encourage adoption of differentiable rendering for challenging, GT-free vector reconstruction tasks across the remote sensing domain. Our key contributions are:
\begin{itemize}
    \item We reframe road network extraction as a global optimization problem over a parametric, curve-based graph. To our knowledge, our work is the first to enable the end-to-end optimization of such a representation directly from segmentation masks, without vector or topology GT.
    \item We propose the B\'{e}zier Graph, a curve-based representation for road networks that is compact, inherently smooth, and fully differentiable, providing the foundation for our optimization-based approach.
    \item We introduce \textit{DOGE}, a differentiable framework that decouples 
    geometric and topological optimization via two complementary modules: \textit{DiffAlign} for differentiable geometry alignment and \textit{TopoAdapt} for discrete topology refinement.
\end{itemize}

\section{Related Work}
\label{sec:related_work}
\subsection{Polyline-based Road Network Modeling}
\label{subsec:polyline-level}
Prior work leveraging polyline representations can be grouped into three main paradigms based on how they construct the graph.

\textbf{Detection and Connection.} These methods first identify road and junctions, then infer connectivity to assemble the graph in a two-stage process~\cite{bahlSingleShotEndtoendRoad2022,heTDRoadTopDownRoad2022,wangRegularizedPrimitiveGraph2023a,shitRelationformerUnifiedFramework2022,hetangSegmentAnythingModel2024,Yin_TowardsSatelliteImage_2025,zaoTopologyGuidedRoadGraph2024a}. While their specifics vary from node extraction (dense detection, heatmaps, 
segmentation) to edge inference (pairwise classification, orientation cues, or learned relational 
reasoning), they all rely on an explicit node-edge decomposition.

\textbf{Iterative Growth.} In contrast, iterative methods reconstruct the network via a sequential decision process, growing the graph from seed points~\cite{bastaniRoadTracerAutomaticExtraction2018,Chu_2019_ICCV_NTG,tanVecRoadPointBasedIterative2020,xuRNGDetRoadNetwork2022,xuRNGDetRoadNetwork2023}. RoadTracer~\cite{bastaniRoadTracerAutomaticExtraction2018} is a canonical example, using a CNN to guide expansion. While effective at capturing local dependencies, these agent-based models can accumulate errors and are inherently local, as previously generated sections of the graph remain fixed.

\textbf{One-Shot Reconstruction.} This paradigm infers the entire graph in a single forward pass, either by decoding compact topological tensors~\cite{heSat2GraphRoadGraph2020,zaoRoadGraphExtraction2024} or by predicting vectorized primitives that are then assembled~\cite{bahlSingleShotEndtoendRoad2022,Xu_PatchedLineSegment_2024,huPolyRoadPolylineTransformer2024}. 
For instance, PaLiS~\cite{Xu_PatchedLineSegment_2024} predicts patch-wise line segments, while PolyRoad~\cite{huPolyRoadPolylineTransformer2024} generates polyline road instances with a transformer.

While effective, the polyline approximation often limits geometric fidelity and editability, and often rely on post-processing for vectorization. 
These limitations motivate our exploration of curve-based representation, paired with a global dynamic optimization strategy.

\subsection{Curve-based Road Network Modeling}
\label{subsec:parametric_curves}
The use of parametric curves for end-to-end SD road network extraction from satellite imagery remains largely unexplored. 
We therefore turn to the related, finer-grained domain of High-Definition (HD) mapping for autonomous driving, where parametric curves like B\'{e}zier curves and splines are well-established for modeling precise lane-level geometry~\cite{chen2022road, liHDMapNetOnlineHD2022, liu2023MapTR, lu2025deeproad}. Prior work in this domain has demonstrated various strategies: directly regressing curve control points~\cite{fengRethinkingEfficientLane2022}, predicting piecewise segments~\cite{qiaoEndtoEndVectorizedHDmap2023}, unifying detection across 2D and 3D modalities~\cite{Dong2024BzierFormerAU}, and, more recently, forming complete lane graphs with explicit topological constraints~\cite{blayneyBezierEverywhereAll2024,liaoLaneGraphPath2025a}.

For instance, the lane-level B\'{e}zier graphs in~\cite{blayneyBezierEverywhereAll2024} enforce geometric smoothness by binding edge tangents to node-shared directions. In contrast, our SD-map formulation offers greater flexibility by optimizing per-edge offsets for internal control points, as detailed in \cref{subsec:bezier_graph}.

However, these HD mapping techniques are ill-suited for our task. They operate at lane level in relatively constrained driving scenes, whereas we aim to reconstruct topologically complete SD road networks at city scale from satellite imagery. Moreover, this entire line of work fundamentally relies on dense vector GT annotations for supervision~\cite{fengRethinkingEfficientLane2022,qiaoEndtoEndVectorizedHDmap2023,Dong2024BzierFormerAU,blayneyBezierEverywhereAll2024,liu2022vectormapnet}, which are subjective and prohibitively expensive to obtain at satellite scale~\cite{lu2025deeproad}. In contrast, we sidestep this dependency by reframing the problem as a global optimization over a curve-based road graph directly supervised by segmentation masks.

\subsection{Differentiable Vector Graphics Rasterization}
\label{subsec:diff_rendering}

Differentiable rasterization allows gradients to flow from a pixel-based loss back to the parameters of vector graphics, such as the control points and stroke widths of B\'{e}zier curves. This is typically achieved by approximating the non-differentiable, hard-edged pixel boundaries of traditional rasterization with a smooth, differentiable function that measures coverage. The canonical implementation, diffvg~\cite{liDifferentiableVectorGraphics2020a}, provides stable gradients that enable optimizing vector shapes to match a target image.

This technique has proven effective for learning compact vector representations in various domains. Applications include layer-wise image vectorization~\cite{xu2022live}, vector font synthesis~\cite{wang2021deepvecfont, wang2023deepvecfont}, and generative art where B\'{e}zier strokes are optimized to create sketches or SVGs guided by semantic models like CLIP or diffusion models~\cite{vinker2022clipasso,Vinker_2023_ICCV,xing2024svgdreamer,xing2023diffsketcher}.


In contrast to prior applications that optimize largely unstructured vector primitives, our work is the first to leverage differentiable rendering for large-scale, topology-aware road network extraction, coupling a structured curve-graph representation with a dynamic optimization process that jointly enforces geometric fidelity and topological correctness without requiring vector GT.
\section{Method}
\label{sec:method}
This section details our framework, \textit{DOGE}, which reconstructs road networks by optimizing a B\'{e}zier Graph (\cref{subsec:bezier_graph}). Our approach decouples this task into two complementary modules: \textit{DiffAlign} (\cref{subsec:diffalign}), which uses differentiable rendering for continuous geometric optimization, and \textit{TopoAdapt} (\cref{subsec:topoadapt}), which applies discrete operators for topology refinement. These modules form an optimization loop (\cref{subsec:optimization_workflow}) that holistically refines the graph.

\subsection{Parametric Representation of B\'{e}zier Graph}
\label{subsec:bezier_graph}

\begin{figure}[htbp]
    \centering
    \begin{tikzpicture}
        \node[anchor=south west,inner sep=0] (image) at (0,0) {\includegraphics[width=0.7\linewidth]{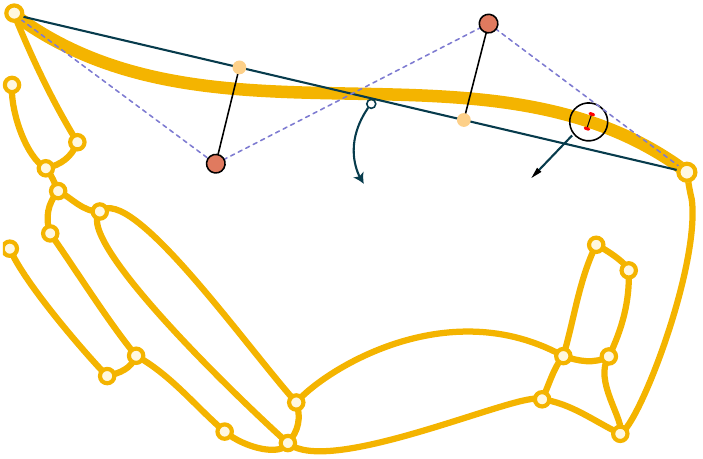}};
        \begin{scope}[x={(image.south east)},y={(image.north west)}]
            \node at (0.70,0.60) {$w_k$};
			\node at (0.55,0.55) {$e_k$};
            \node at (-0.06,0.98) {$\boldsymbol{P}_{k,0}$};
            \node at (0.23,0.94) {$\alpha_0$};
            \node at (0.38,0.58) {$\boldsymbol{P}_{k,1}$};
            \node at (0.79,0.95) {$\boldsymbol{P}_{k,2}$};
            \node at (0.85,0.62) {$\alpha_1$};
            \node at (1.00,0.70) {$\boldsymbol{P}_{k,3}$};
        \end{scope}
    \end{tikzpicture}
    \caption{Parametric definition of a B\'{e}zier Graph edge $e_k$. The edge's geometry is defined by a cubic B\'{e}zier curve with four control points, $\{\boldsymbol{P}_{k,r}\}_{r=0}^3$, and an optimizable width, $w_k$. The endpoints $\boldsymbol{P}_{k,0}$ and $\boldsymbol{P}_{k,3}$ are anchored to the node positions, while the intermediate points $\boldsymbol{P}_{k,1}$ and $\boldsymbol{P}_{k,2}$ control the curvature.}
    \label{fig:bezier_graph}
    \vspace{-1.5mm}
\end{figure}
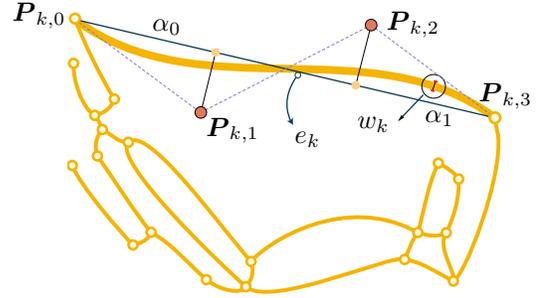

We formally model the road layout as a B\'{e}zier Graph $\mathcal{G} = (\mathcal{V}, \mathcal{E})$. The node set, $\mathcal{V} = \{v^k\}$, consists of vertices with optimizable 2D positions, $\mathbf{p}_k \in \mathbb{R}^2$, representing intersections, termini, or points along a road's path based on their degree. The edge set, $\mathcal{E} = \{e_k\}$, connects these nodes with curvilinear road segments.

As illustrated in \cref{fig:bezier_graph}, each edge's geometry is given by a cubic B\'{e}zier curve:
\begin{equation}
\boldsymbol{C}_k(t) = \sum_{r=0}^{3} \binom{3}{r} (1-t)^{3-r} t^r \boldsymbol{P}_{k,r}, \quad t \in [0, 1].
\label{eq:bezier_curve}
\end{equation}
To ensure a regularized and well-posed representation amenable to optimization, we do not optimize its four control points $\{\boldsymbol{P}_{k,r}\}_{r=0}^3$ directly. Instead, they are deterministically constructed as follows:
\begin{subequations}
\label{eq:control_points}
\begin{align}
    \boldsymbol{P}_{k,0} &= \mathbf{p}_i \\
    \boldsymbol{P}_{k,3} &= \mathbf{p}_j \\
    \boldsymbol{P}_{k,1} &= \left((1-\alpha_{k,0})\boldsymbol{P}_{k,0} + \alpha_{k,0}\boldsymbol{P}_{k,3}\right) + d_{k,0} \cdot \mathbf{n}_{ij} \label{eq:p1}\\
    \boldsymbol{P}_{k,2} &= \left((1-\alpha_{k,1})\boldsymbol{P}_{k,0} + \alpha_{k,1}\boldsymbol{P}_{k,3}\right) + d_{k,1} \cdot \mathbf{n}_{ij} \label{eq:p2}
\end{align}
\end{subequations}
While the endpoints $\boldsymbol{P}_{k,0}$ and $\boldsymbol{P}_{k,3}$ are anchored to their corresponding node positions to ensure topological connectivity, we reparameterize the intermediate points $\boldsymbol{P}_{k,1}$ and $\boldsymbol{P}_{k,2}$ to reduce the degrees of freedom. As detailed in Eq.~\ref{eq:control_points}, their positions are defined by two learnable scalars each: a projection parameter $\alpha_{k,i} \in [0, 1]$ that places a point along the chord $\overline{\boldsymbol{P}_{k,0} \boldsymbol{P}_{k,3}}$, and an offset distance $d_{k,i} \in \mathbb{R}$ that displaces it perpendicularly. This reparameterization regularizes the curve's shape—preventing degeneracies like self-intersections—while maintaining sufficient flexibility. The normal vector $\mathbf{n}_{ij}$ is the normalized perpendicular of the chord vector.

\definecolor{customOrange}{HTML}{e97132}
\definecolor{customBlueGray}{HTML}{afc3e4}
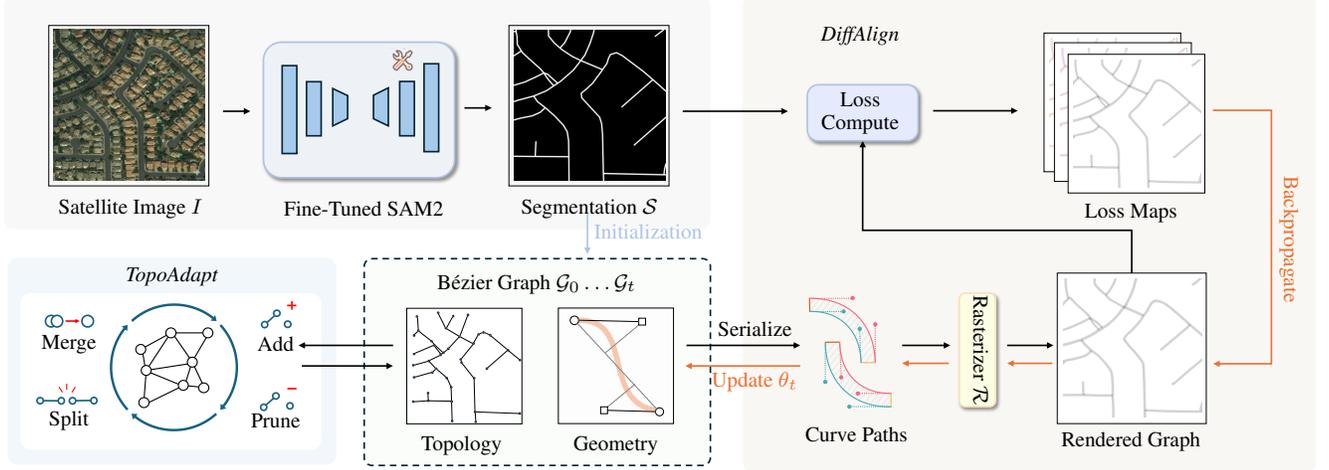
\begin{figure*}[htbp]
    \centering
    \begin{tikzpicture}
        \node[anchor=south west,inner sep=0] (image) at (0,0) {\includegraphics[width=\textwidth]{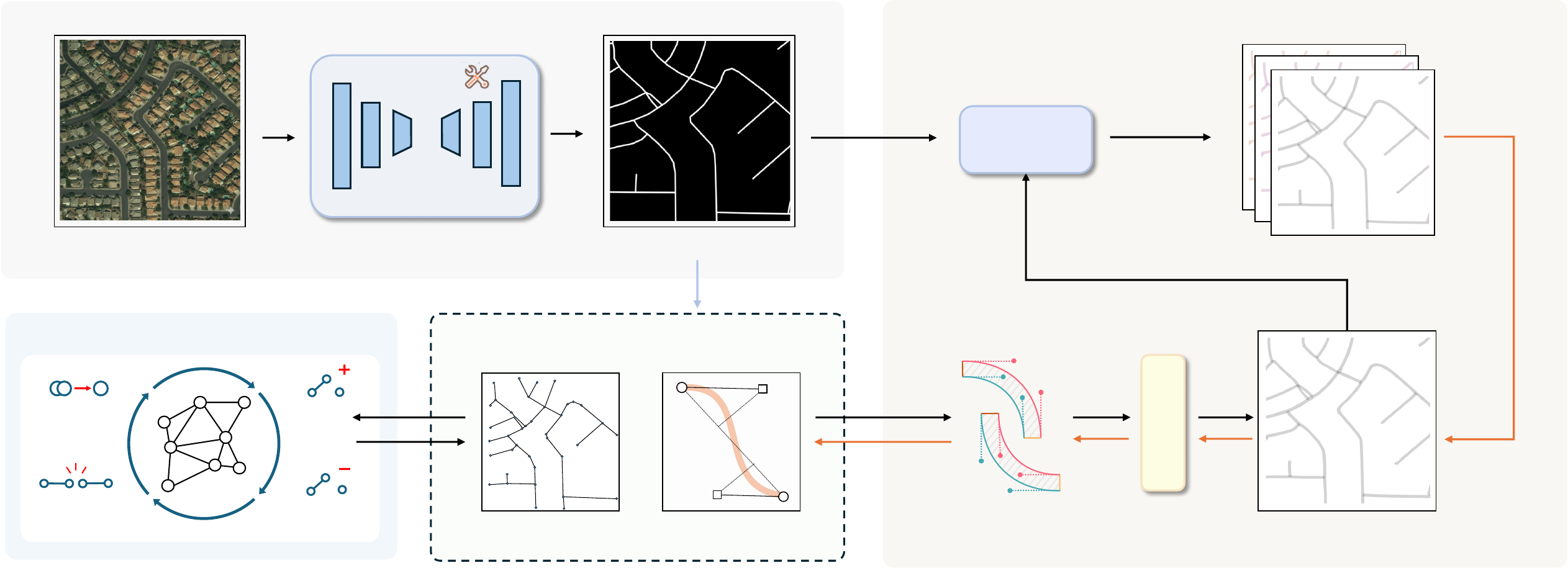}};
        
        \begin{scope}[every node/.style={font=\footnotesize}]
            \node at (1.7, 3.5) {Satellite Image $I$};
            \node at (4.8, 3.5) {Fine-Tuned SAM2};
            \node at (7.8, 3.5) {Segmentation $\mathcal{S}$};

            \node at (7.1, 2.5) {B\'{e}zier Graph $\mathcal{G}_0 \dots \mathcal{G}_t$};
            \node[text=customBlueGray] at (8.58, 3.2) {Initialization};

            \node at (2.25, 2.60) {\textit{TopoAdapt}};
            \node at (6.1, 0.35) {Topology};
            \node at (3.63, 1.7) {Add};
            \node at (0.88, 1.7) {Merge};
            \node at (0.88, 0.68) {Split};
            \node at (3.63, 0.68) {Prune};

            \node at (11.4, 5.8) {\textit{DiffAlign}};
            \node at (8.15, 0.35) {Geometry};
            
            \node[rotate=-90] at (13, 1.62) {Rasterizer $\mathcal{R}$};
            \node at (10, 1.90) {Serialize};
            \node[text=customOrange] at (10, 1.2) {Update $\theta_t$};
            \node at (11.35, 0.5) {Curve Paths};
            
            \node[rotate=-90, text=customOrange] at (17.1, 3.1) {Backpropagate};
            
            \node at (11.4, 4.63) {Compute};
            \node at (11.4, 4.95) {Loss};

            \node at (15, 3.45) {Loss Maps};
            \node at (15, 0.41) {Rendered Graph};

        \end{scope}
    \end{tikzpicture}
    \vspace{-6mm}
    \caption{Overview of the \textit{DOGE} framework. Given a satellite image, a fine-tuned SAM2 provides a target road segmentation $\mathcal{S}$. \textit{DOGE} reconstructs the road network by iteratively optimizing a B\'{e}zier Graph $\mathcal{G}$ (\cref{subsec:bezier_graph}). The optimization loop alternates between two complementary modules: \textit{DiffAlign}, which continuously refines the graph's geometry by aligning a differentiable rendering of the graph with $\mathcal{S}$ (\cref{subsec:diffalign}), and \textit{TopoAdapt}, which discretely evolves the graph's topology (\cref{subsec:topoadapt}).}
    \vspace{-3mm}
    \label{fig:overview}
\end{figure*}

\subsection{DiffAlign: Differentiable Geometric Optimization}
\label{subsec:diffalign}

The geometric optimization is guided by a target segmentation $\mathcal{S}$, obtained from a fine-tuned SAM2 model. As illustrated in \cref{fig:overview}, the process begins by serializing each B\'{e}zier curve into a closed polygon representing a road segment. Further details on this serialization are provided in the supplementary material. These polygons are then passed to a differentiable rasterizer $\mathcal{R}$ (DiffVG~\cite{liDifferentiableVectorGraphics2020a}) to produce a rendered map of the road network. We then optimize the B\'{e}zier Graph by computing losses on this rendered output and applying direct, vector-space regularization between curves. This process refines the graph's geometric parameters $\theta$, which include the node positions $\{\mathbf{p}_k\}$, edge widths $\{w_k\}$, and the reparameterized curve attributes $\{\alpha_{k,i}, d_{k,i}\}$.

\paragraph{Objective Function}
The optimization is driven by a composite objective, $\mathcal{L}_{\text{total}}$, which combines five weighted loss terms. These terms are grouped into a target-alignment loss for data fidelity and four geometric priors that regularize the graph's structure. The total loss is:
\begin{equation}
\label{eq:geom_loss}
\begin{split}\mathcal{L}_{\text{total}} ={}& \lambda_{\text{cover}} \mathcal{L}_{\text{cover}} + \lambda_{\text{overlap}} \mathcal{L}_{\text{overlap}} + \lambda_{G1} \mathcal{L}_{G1} \\
    & + \lambda_{\text{offset}} \mathcal{L}_{\text{offset}} + \lambda_{\text{spacing}} \mathcal{L}_{\text{spacing}}.
\end{split}
\end{equation}
The $\lambda$ terms are weighting coefficients of each loss components. The specific values are detailed in \cref{subsec:implementation_details}.

\paragraph{Target–Image Alignment}
The \textbf{Coverage Loss} is our data-fidelity term. It penalizes the pixel-wise L2 discrepancy between the target segmentation $\mathcal{S}$ and the union of rendered edges, ensuring the graph covers the road regions:
\begin{equation}
	\mathcal{L}_{\text{cover}} = \left\lVert \left(\bigcup_{e_k \in \mathcal{E}_t} R(e_k)\right) - \mathcal{S} \right\rVert_2^2
\label{eq:coverage_loss}
\end{equation}

\paragraph{Geometric and Topological Priors}
To guide the optimization towards a plausible road network, we introduce four regularization priors. These terms operate independently of $\mathcal{S}$ and enforce desirable geometric properties.

First, the \textbf{Overlap Loss} promotes correct topology by penalizing the total area of improper intersections between rendered road segments. To compute this, an overlap map is formed by summing all individual edge renderings and clipping the result at a value of 1. The loss is the L1 norm of this map, normalized by the number of edges $N_t$:
\begin{equation}
	\mathcal{L}_{\text{overlap}} = \frac{1}{N_t} \left\lVert \max\left(0, \left(\sum_{e_k \in \mathcal{E}_t} R(e_k)\right) - 1 \right) \right\rVert_1
	\label{eq:overlap_loss}
\end{equation}

Second, the \textbf{G1 Continuity Loss} encourages tangent alignment at degree-2 nodes. For a node $v_j$ connecting edges $e_a$ and $e_b$, the tangents are defined by vectors $\mathbf{v}_{\text{in}} = \mathbf{p}_j - \boldsymbol{P}_{a,2}$ and $\mathbf{v}_{\text{out}} = \boldsymbol{P}_{b,1} - \mathbf{p}_j$. The loss is:
\begin{equation}
\mathcal{L}_{G1} = \frac {1}{N_t}\sum_{\substack{v_j \in \mathcal{V}_t \\ \text{deg}(v_j)=2}} (1 - \cos(\theta_j)) \cdot \mathbf{1}_{(\theta_j < T_{G1})}
\end{equation}
where $\theta_j$ is the angle between tangents at node $v_j$. The loss activates only when $\theta_j$ falls below a threshold $T_{G1}$, penalizing nearly straight connections to preserve legitimate turns.

Finally, two curve regularization terms penalize ill-formed B\'{e}zier geometries. The \textbf{Offset Loss} discourages excessive perpendicular offsets of the intermediate control points, while the \textbf{Spacing Loss} encourages their projection parameters, $\alpha_{k,0}$ and $\alpha_{k,1}$, to be evenly spaced along the chord connecting the endpoints.

\begin{equation}
    \label{eq:offset_loss}
    \begin{split}
        \mathcal{L}_{\text{offset}} = \frac{1}{N_t} \sum_{\substack{e_k \in \mathcal{E}_t \\ i \in \{0, 1\}}} \max\Big(0, &\exp\Big(\frac{|d_{k,i}|}{L_k} - \tau_d\Big) - 1\Big) \raisetag{8pt}
    \end{split}
\end{equation}

\begin{equation}
\mathcal{L}_{\text{spacing}} = \frac{1}{N_t} \sum_{e_k \in \mathcal{E}_t} \left( (\alpha_{k,0} - \hat \alpha_{0})^2 + (\alpha_{k,1} - \hat \alpha_{1})^2 \right)
\label{eq:spacing_loss}
\end{equation}
Here, $L_k = \|\boldsymbol{P}_{k,3} - \boldsymbol{P}_{k,0}\|$ is the chord length of edge $e_k$, $\tau_d$ is a threshold for the maximum allowable offset ratio, and the constants $\hat \alpha_{0}, \hat \alpha_{1}$ are set to 1/3 and 2/3 respectively, representing the ideal equidistant positions for the control points along the chord.

\subsection{TopoAdapt: Discrete Topology Refinement}
\label{subsec:topoadapt}

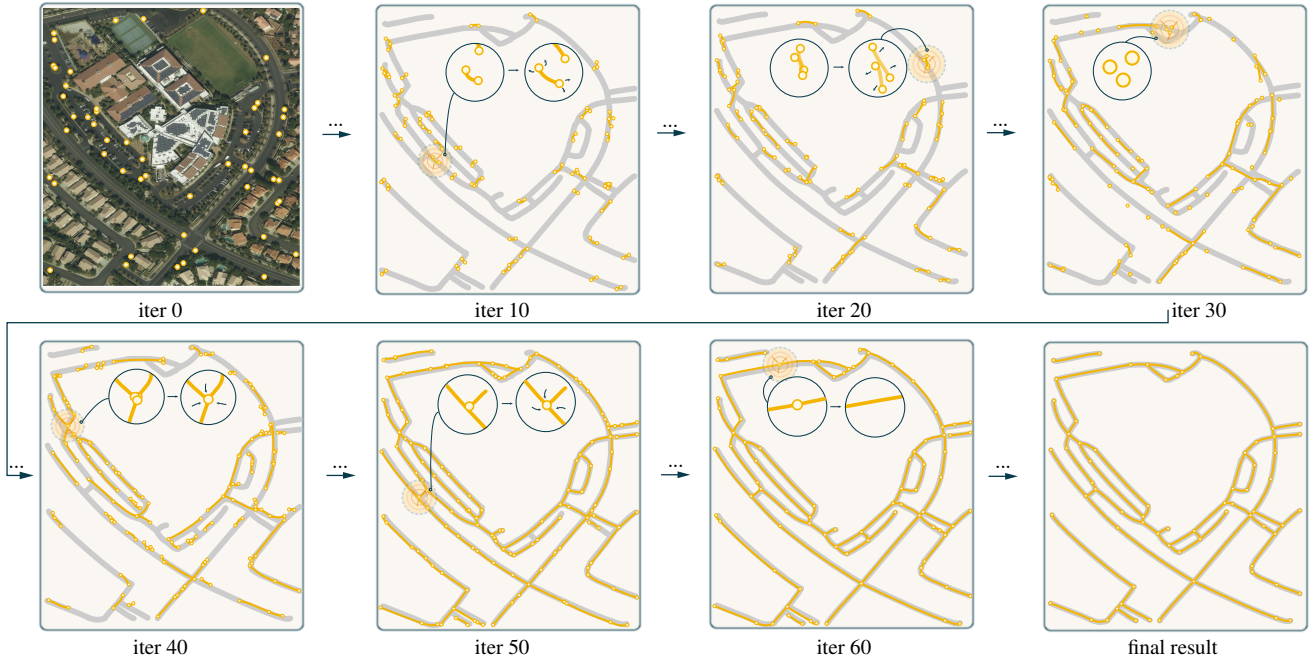
\begin{figure*}[!t]
    \centering

	\begin{tikzpicture}
        \node[anchor=south west,inner sep=0] (image) at (0,0) {\includegraphics[width=\textwidth]{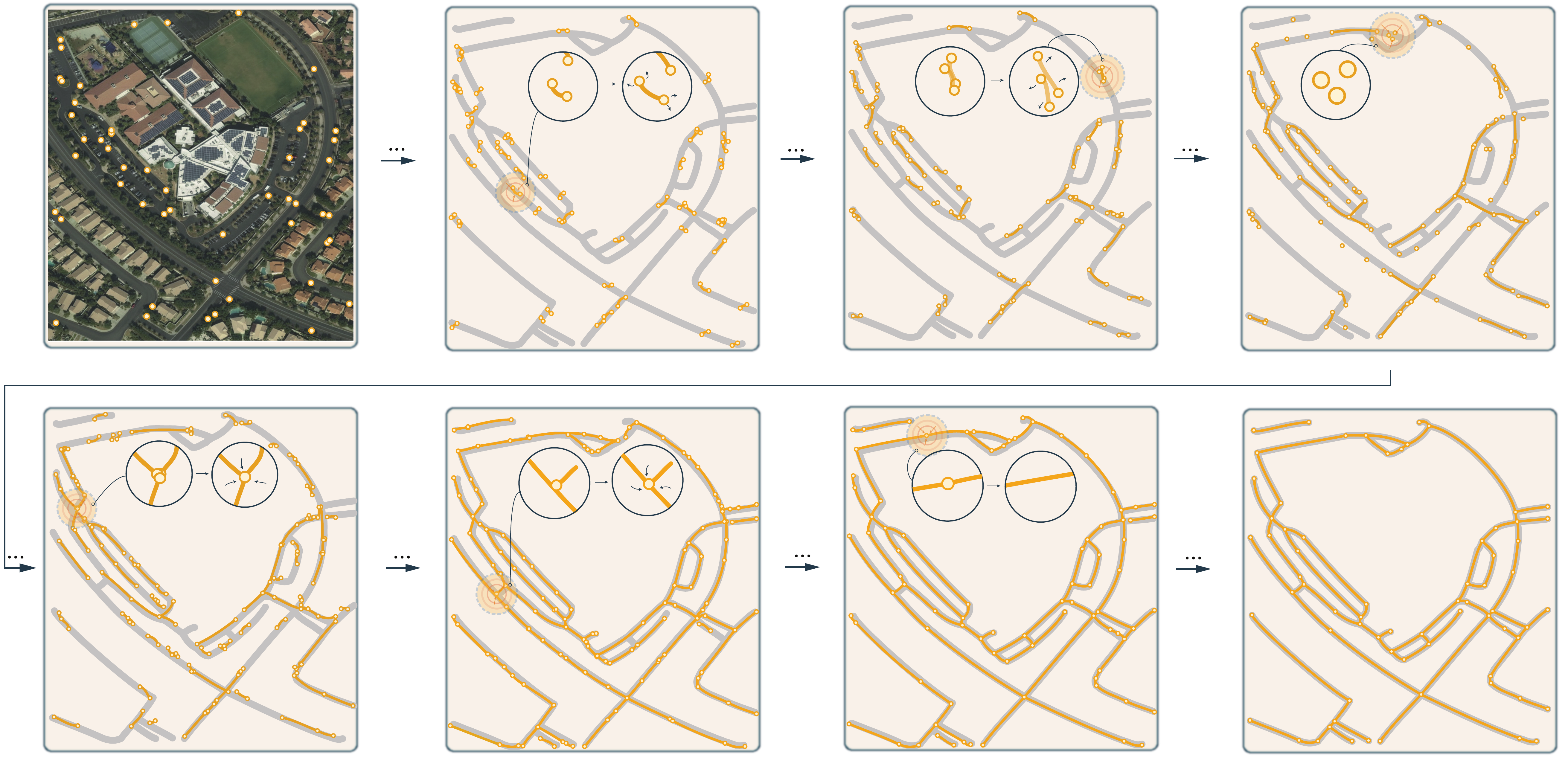}};
        \begin{scope}[x={(image.south east)},y={(image.north west)}]

			\node at (0.12,0.52) {\footnotesize iter 0};
			\node at (0.38,0.52) {\footnotesize iter 10};
			\node at (0.64,0.52) {\footnotesize iter 20};
			\node at (0.91,0.52) {\footnotesize iter 30};

			\node at (0.12,0.00) {\footnotesize iter 40};
			\node at (0.38,0.00) {\footnotesize iter 50};
			\node at (0.64,0.00) {\footnotesize iter 60};
			\node at (0.89,0.00) {\footnotesize final result};
        \end{scope}
    \end{tikzpicture}

    \vspace{-1.5mm}
    \caption{Optimization dynamics of the B\'{e}zier Graph. This figure illustrates the interplay between \textit{DiffAlign} and \textit{TopoAdapt}. Key operations are highlighted: graph initialization (iter 0); geometric optimization towards the target (iter 10); overlap separation driven by $\mathcal{L}_{\text{overlap}}$ (iter 20); road addition (iter 30); node merging (iter 40); T-junction creation (iter 50); and collinear edge merging (iter 60).}
    \label{fig:optimization_dynamics}
    \vspace{-2mm}
\end{figure*}

While \textit{DiffAlign} continuously refines geometry, \textit{TopoAdapt} complements by applying a set of discrete, heuristic operators to dynamically refine the graph's topology. Notably, these operators rely on a handful of simple, robust thresholds that are kept fixed across all datasets. In \cref{fig:optimization_dynamics}, this allows for corrections like adding missing roads or merging redundant nodes. These operators, accelerated by a spatial grid for efficient querying, fall into three categories.

\textbf{Road Addition}
To grow the graph, we identify regions where the current graph $\mathcal{G}_t$ inadequately covers the target segmentation $\mathcal{S}$ by computing a difference map using the current rendered graph:
\begin{equation}
M_{\text{unfit}} = \mathbf{1}[\mathcal{S} > \tau_{\text{seg}}] \odot \mathbf{1}[\mathrm{Render}(\mathcal{G}_t) < \tau_{\text{render}}].
\end{equation}
From this map, we sample $k$ candidate locations $\{p_i\}$ to instantiate new road segments. For each location $p_i$, a new edge is created by initializing two endpoints, $v'_{i,0}$ and $v'_{i,1}$, at positions randomly offset from the center (e.g., $\mathbf{p}'_{i,0/1} = p_i \pm \frac{L}{2}\mathbf{u}_i$ for a small length $L$ and a random unit vector $\mathbf{u}_i$). The connecting B\'{e}zier edge is initialized with small, random scalar offsets ($d_{i,0}, d_{i,1}$) to form a nearly straight line. The new nodes and edges are then added to the graph:
\begin{equation}
\mathcal{G}_{t} \leftarrow (\mathcal{V}_t \cup \mathcal{V}_{\text{new}}, \mathcal{E}_t \cup \mathcal{E}_{\text{new}}).
\end{equation}

\textbf{Connectivity Enhancement}
We connect nearby graph components with two operators based on proximity.
\begin{itemize}
    \item \textbf{Node Merging:} Pairs of nodes $(v_i, v_j)$ closer than a distance $\epsilon_{\text{merge}}$ are merged into a single new node at their midpoint, which inherits all incident edges.
    \item \textbf{T-Junction Creation:} A node $v_i$ is snapped to a nearby edge $e_j$ if their minimum distance is less than $\epsilon_{\text{merge}}$. This is achieved by splitting $e_j$ at the closest point to $v_i$ and merging the new vertex with $v_i$.
\end{itemize}

\textbf{Graph Simplification and Pruning}
To maintain a clean and efficient representation, we periodically apply simplification operators.
\begin{itemize}
    \item \textbf{Collinear Edge Merging:} Degree-2 nodes that lie on nearly straight paths are removed, and their two incident edges are replaced by a single, refitted B\'{e}zier curve. A path is deemed straight if the angle between its edge tangents exceeds a threshold.
    \item \textbf{Invalid Edge Pruning:} Geometrically implausible edges, such as those that are too short or too thin, are pruned. Any resulting isolated (degree-0) nodes are also removed.
\end{itemize}

\subsection{Global Dynamic Optimization}
\label{subsec:optimization_workflow}

Our framework uses global dynamic optimization to holistically refine the road network. As summarized in \cref{alg:main_algorithm}, this is achieved via a loop that alternates between \textit{TopoAdapt} and \textit{DiffAlign}. Concretely, we initialize the graph $\mathcal{G}_0$ by applying the Road Addition procedure on the segmentation $\mathcal{S}$ to obtain an initial set of B\'{e}zier edges covering high-confidence road regions. In each iteration, \textit{TopoAdapt} refines the graph's connectivity ($\mathcal{V}, \mathcal{E}$) before \textit{DiffAlign} optimizes its geometric parameters ($\theta$). Detailed operator algorithms are in the supplementary material.

\vspace{-4pt}
\begin{algorithm}[h]
    \caption{Optimization Workflow of \textit{DOGE}}
    \label{alg:main_algorithm}
    \KwIn{Target segmentation $\mathcal{S}$, Max iterations $T_{max}$, Learning rate $\eta$}
    \KwOut{Optimized B\'{e}zier Graph $\mathcal{G}^* = (\mathcal{V}^*, \mathcal{E}^*, \theta^*)$}
    
    $\mathcal{G}_0 = (\mathcal{V}_0, \mathcal{E}_0, \theta_0) \leftarrow \text{InitializeGraph()}$;
    
    \For{$t = 0$ \KwTo $T_{max}-1$}
    {
        $(\mathcal{V}'_t, \mathcal{E}'_t) \leftarrow \text{\textit{TopoAdapt}}(\mathcal{V}_t, \mathcal{E}_t)$; 
        $\mathcal{G}'_t \leftarrow (\mathcal{V}'_t, \mathcal{E}'_t, \theta_t)$;
        
        $\theta_{t+1} \leftarrow \text{\textit{DiffAlign}}(\mathcal{G}'_t, \mathcal{S}, \eta)$;
        
        $\mathcal{G}_{t+1} \leftarrow (\mathcal{V}'_t, \mathcal{E}'_t, \theta_{t+1})$;
    }
    
    \Return{$\mathcal{G}_{T_{max}}$};
\end{algorithm}

\section{Experiments}
\label{sec:experiments}

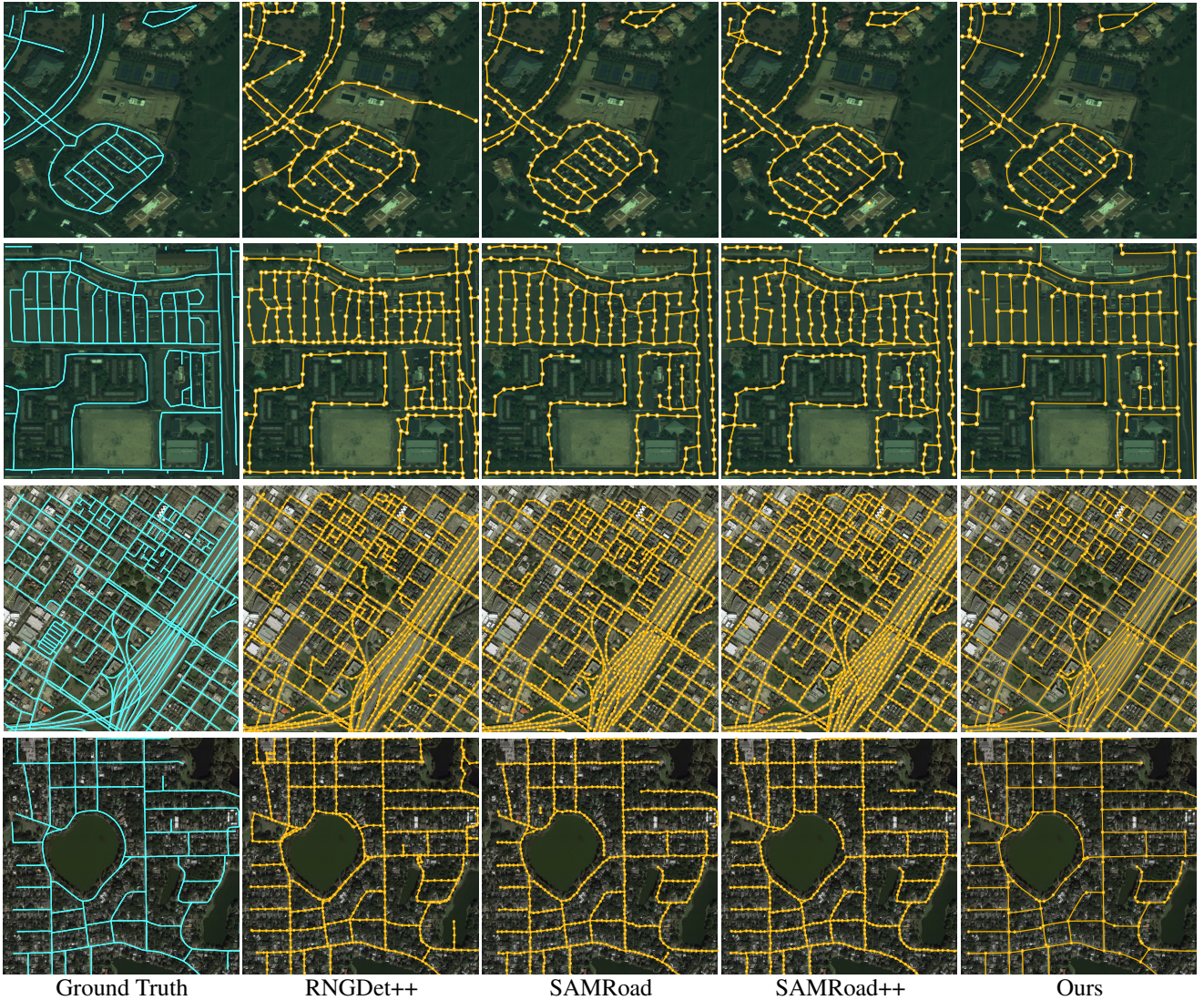
\begin{figure*}[!t]
    \centering
    \begin{tikzpicture}
        \node[anchor=south west,inner sep=0] (image) at (0,0) {\includegraphics[width=\linewidth]{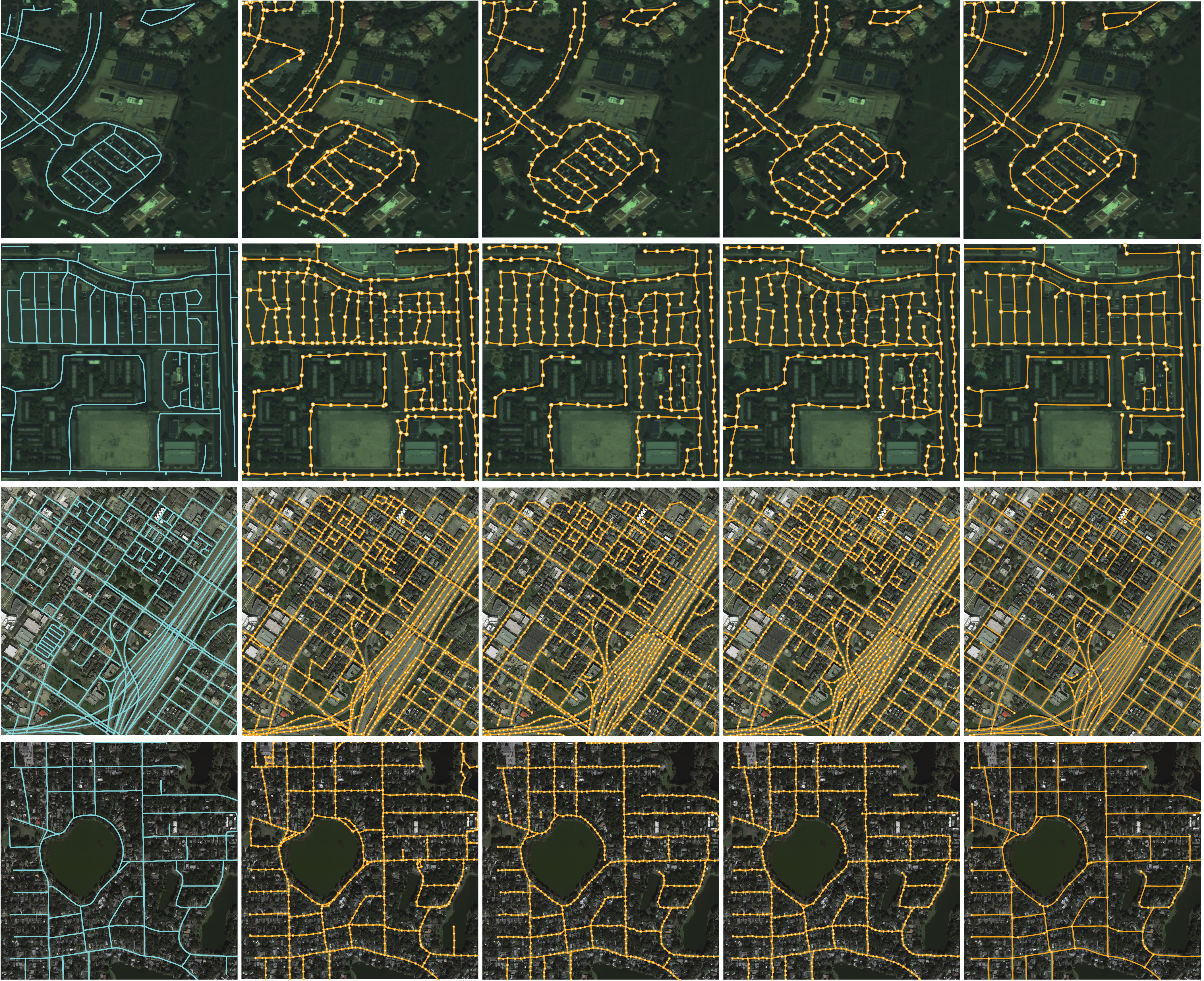}};
        \begin{scope}[x={(image.south east)},y={(image.north west)}]
            \node at (0.1, -0.012) {Ground Truth};
            \node at (0.3, -0.012) {RNGDet++};
            \node at (0.5, -0.012) {SAMRoad};
            \node at (0.7, -0.012) {SAMRoad++};
            \node at (0.9, -0.012) {Ours};
        \end{scope}
    \end{tikzpicture}
    \caption{Qualitative comparison on SpaceNet (top two rows) and City-Scale (bottom two rows). Our method produces geometrically precise, smooth, and topologically correct road graphs, outperforming prior methods across different scales. Notably, our approach uses a more compact graph representation with fewer nodes.}
    \label{fig:qualitative_comparison}
\end{figure*}


We evaluate on two public benchmarks: City-Scale (Sat2Graph) and SpaceNet. Unless otherwise stated, all images are standardized to 1 m/pixel following prior work, consistent with SAMRoad and SAMRoad++~\cite{hetangSegmentAnythingModel2024,Yin_TowardsSatelliteImage_2025}.

\textbf{City-Scale (Sat2Graph).} City-Scale (Sat2Graph)~\cite{heSat2GraphRoadGraph2020} contains 180 RGB satellite tiles at $2048\times 2048$ covering multiple U.S. cities, with road networks provided as vector graphs. We adopt the common split $144/9/27$ for train/val/test, identical to Sat2Graph, RNGDet++, and SAMRoad for fair comparison.

\textbf{SpaceNet.} SpaceNet~\cite{Etten2018SpaceNetAR} comprises roughly 2.5k RGB tiles at $400\times 400$ from diverse world cities, with vector road-graph annotations. We follow prior work (Sat2Graph/RNGDet++/SAMRoad) and use the $2042/127/382$ train/val/test split. Following SAMRoad/SAMRoad++ preprocessing, images are resampled to 1 m/pixel for consistent resolution.

\begin{table*}[!b]
    \centering
    \caption{Performance comparison with state-of-the-art methods on the SpaceNet and City-Scale datasets. The best results are shown in \textbf{bold}, and the second-best are \underline{underlined}.}
    \label{tab:quantitative_comparison}
    \small
    \begin{tabular}{lcccccccc}
        \toprule
        \multirow{2}{*}{Method} & \multicolumn{4}{c}{SpaceNet} & \multicolumn{4}{c}{City-Scale} \\
        \cmidrule(lr){2-5} \cmidrule(lr){6-9}
        & TOPO F1 $\uparrow$ & Precision $\uparrow$ & Recall $\uparrow$ & APLS $\uparrow$ & TOPO F1 $\uparrow$ & Precision $\uparrow$ & Recall $\uparrow$ & APLS $\uparrow$ \\
        \midrule
        \midrule
        Sat2Graph & 80.97 & 85.93 & \underline{76.55} & 64.43 & 76.26 & 80.70 & 72.28 & 63.14 \\
        RNGDet & 81.13 & 90.91 & 73.25 & 65.61 & 76.87 & 85.97 & 69.87 & 65.75 \\
        RNGDet++ & \underline{82.81} & 91.34 & 75.24 & 67.73 & 78.44 & 85.65 & 72.58 & 67.76 \\
        SAMRoad & 80.52 & 93.03 & 70.97 & 71.64 & 77.23 & \textbf{90.47} & 67.69 & \underline{68.37} \\
        SAMRoad++ & 81.57 & \textbf{93.68} & 72.23 & \underline{73.44} & \underline{80.01} & \underline{88.39} & \underline{73.39} & 68.34 \\
        \midrule
        \textbf{DOGE} & \textbf{84.58} & \underline{93.55} & \textbf{78.43} & \textbf{73.48} & \textbf{80.59} & 84.42 & \textbf{77.40} & \textbf{70.24} \\
        \bottomrule
        \bottomrule
    \end{tabular}
\end{table*}

\subsection{Implementation Details}
\label{subsec:implementation_details}

Our framework is implemented in PyTorch, with differentiable rendering built upon DiffVG~\cite{liDifferentiableVectorGraphics2020a}. The optimization is conducted on a single NVIDIA RTX 4090 GPU.

\textbf{Target Road Segmentation.} We use a fine-tuned SAM2 model to generate target road segmentation masks. The model is trained on the respective training splits of each dataset. Further details on the segmentation model's architecture and training are in the supplementary material.

\textbf{Optimization.} We use the Adam optimizer for geometric parameters. The optimization runs for a maximum of $T_{\max} = 300$ iterations, although we employ an early stopping strategy detailed in the supplementary material. Loss weights in Eq.~\ref{eq:geom_loss} are: $\lambda_{\text{cover}}=1.0$, $\lambda_{\text{overlap}}=0.3$, $\lambda_{G1}=0.012$, and $\lambda_{\text{offset}}=\lambda_{\text{spacing}}=6 \times 10^{-3}$. The rendering resolution is $512 \times 512$. For \textit{TopoAdapt}, we use a unified proximity threshold of $\epsilon_{\text{merge}} = 4\text{m}$ for both node merging and T-junction creation. All \textit{TopoAdapt} hyperparameters are kept fixed across both datasets, underscoring the module's robustness. A comprehensive list of all hyperparameters can be found in the supplementary material.

\subsection{Evaluation Metrics}
\label{subsec:evaluation_metrics}
We follow standard practice~\cite{heSat2GraphRoadGraph2020,xuRNGDetRoadNetwork2023,hetangSegmentAnythingModel2024} and evaluate with the TOPO~\cite{biagioni2012inferring} and APLS~\cite{Etten2018SpaceNetAR} metrics. These metrics primarily assess topology but include weak geometric constraints. For fair comparison with prior work, our B\'{e}zier graph is converted to a polyline compatible with the official evaluation scripts, using their public implementations and default parameters. We obtain these polylines by uniformly sampling points along each B\'{e}zier edge, ensuring a consistent discretization across all methods. The conversion is detailed in the supplementary.

\subsection{Comparative Results}
\label{subsec:comparative_results}

We evaluate our method, \textit{DOGE}, against state-of-the-art methods on the SpaceNet and City-Scale benchmarks.
The qualitative results in \cref{fig:qualitative_comparison} highlight the advantages of our approach. \textit{DOGE} produces road networks that are significantly smoother and more geometrically accurate than those from prior methods like RNGDet++ and SAMRoad++. This is a direct benefit of our B\'{e}zier curve representation, which naturally models the curvilinear nature of roads. Furthermore, our adaptive algorithm efficiently partitions road geometry, using more segments for high-curvature areas and fewer for straight sections. This allows \textit{DOGE} to effectively model complex intersections while avoiding the jagged artifacts common in other approaches.

The quantitative results in \cref{tab:quantitative_comparison} confirm the superiority of our approach. \textit{DOGE} sets a new state-of-the-art on both datasets, leading in TOPO F1 on SpaceNet (\textbf{84.58}) and APLS on City-Scale (\textbf{70.24}). This success is driven by our method's ability to substantially improve recall for a more complete reconstruction, while maintaining a high precision that leads to a superior trade-off on both benchmarks. This demonstrates our method's strength in reconstructing both topologically complete and geometrically accurate road networks.  Our full analysis, including cross-dataset tests and experiments, can be found in the supplementary material.

\subsection{Analysis of Topological Compactness}
\label{subsec:compactness_analysis}

\begin{figure}[h]
    \centering
    \includegraphics[width=0.95	\linewidth]{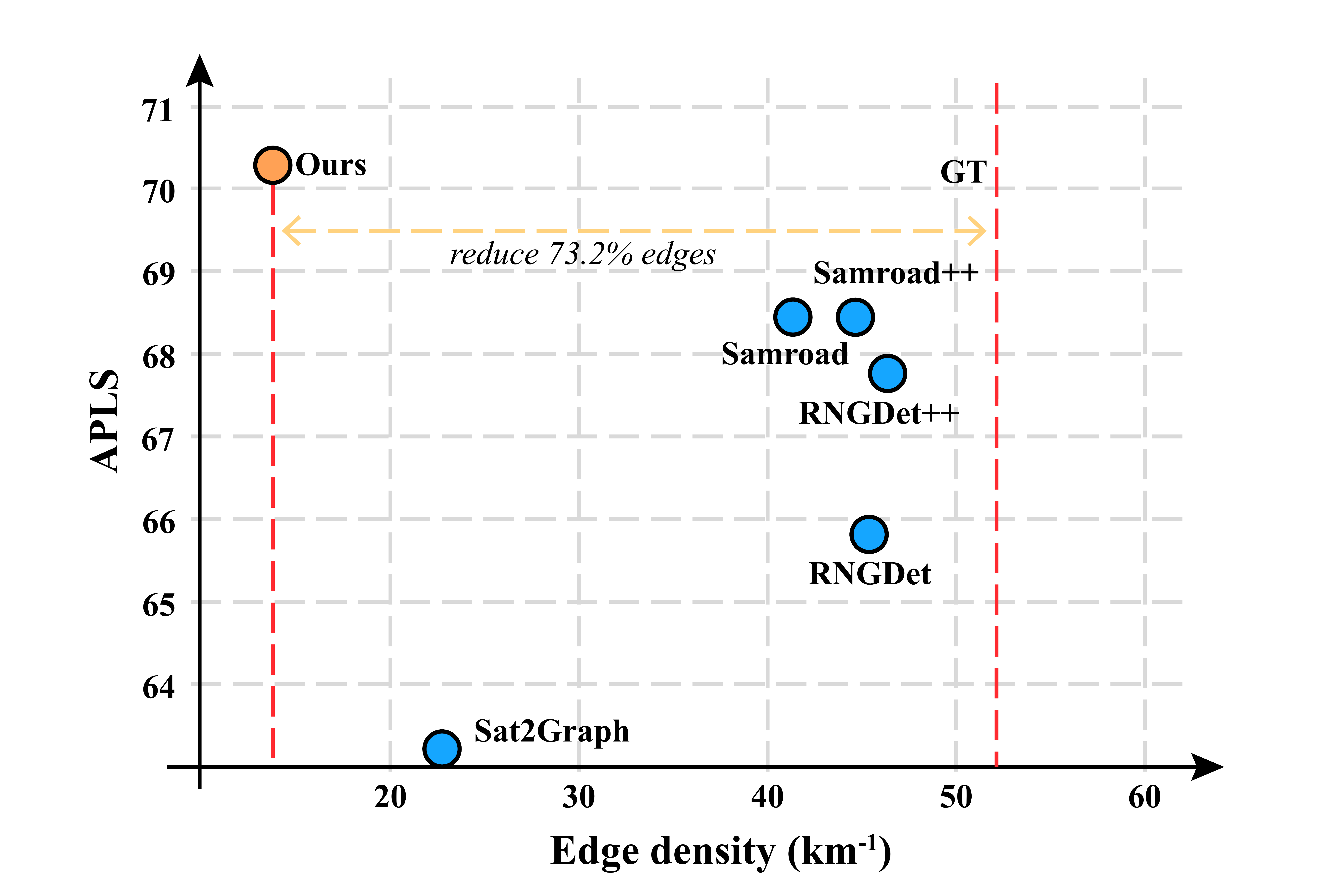}
    \vspace{-1.5mm}
    \caption{Performance versus compactness on the City-Scale dataset. The plot shows APLS against edge density (edges/km). \textit{DOGE}, achieves the best result, delivering the highest APLS score with a significantly more compact graph representation—using 73.2\% fewer edges per kilometer than the GT.}
    \label{fig:edge_density}
    \vspace{-2mm}
\end{figure}

Our method produces road networks that are more accurate and significantly more compact than prior work, a dual advantage evident on the City-Scale dataset. As \cref{fig:edge_density} illustrates, our approach achieves a higher topological accuracy (APLS) with a substantially lower edge density (edges/km). This superior trade-off between fidelity and compactness is a direct result of our global dynamic optimization strategy. \cref{tab:compactness} further quantifies this efficiency, and a full analysis is provided in the supplementary material.

\begin{table}[htbp]
    \centering
    \vspace{-1mm}
    \caption{Compactness and performance of graph representations on the City-Scale test set.}
    \label{tab:compactness}
    \small
    \begin{tabular}{lccc}
        \toprule
        Method & Nodes/km $\downarrow$ & Edges/km $\downarrow$ & APLS $\uparrow$ \\
        \midrule
        \midrule
        GT & 49.36 & 52.17 & -- \\
        Sat2Graph & 18.67 & 22.69 & 63.14 \\
        RNGDet & 42.64 & 45.38 & 65.75 \\
        RNGDet++ & 43.37 & 46.22 & 67.76 \\
        SAMRoad & 39.25 & 41.34 & 68.37 \\
        SAMRoad++ & 41.70 & 44.49 & 68.34 \\
		\midrule
        \textbf{DOGE} & \textbf{11.26} & \textbf{14.00} & \textbf{70.24} \\
        \bottomrule
        \bottomrule
    \end{tabular}
    \vspace{-2mm}
\end{table}


\begin{figure}[t]
    \centering
    \begin{tikzpicture}
        \node[anchor=south west, inner sep=0] (image) at (0,0) {\includegraphics[width=0.90\linewidth]{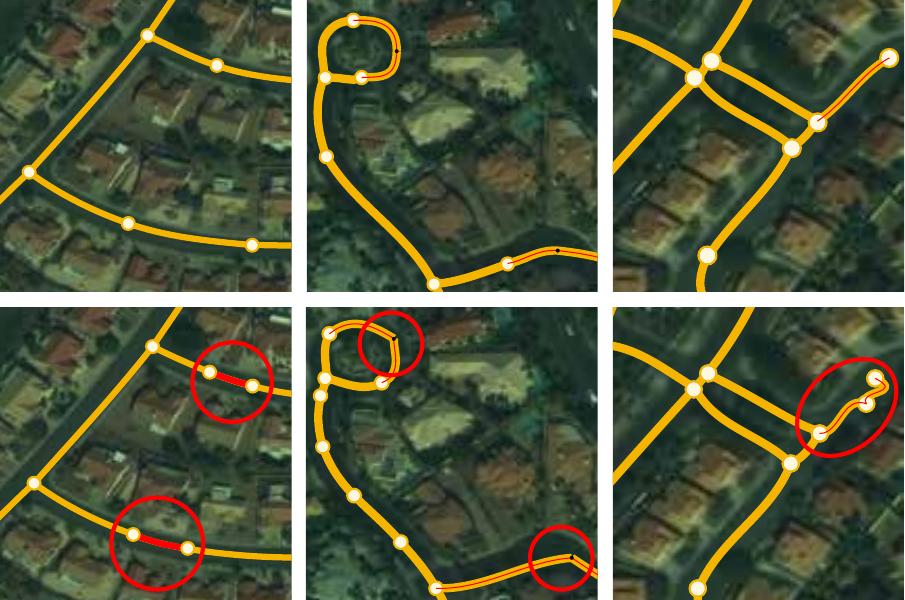}};
        \begin{scope}[x={(image.south east)},y={(image.north west)}]
            \node[font=\footnotesize] at (0.17, -0.04) {(a)};
            \node[font=\footnotesize] at (0.5, -0.04) {(b)};
            \node[font=\footnotesize] at (0.83, -0.04) {(c)};
        \end{scope}
    \end{tikzpicture}
    \vspace{-1.5mm}
    \caption{Effectiveness of our proposed geometric and topological priors. The top row shows the ideal result from our full model. The bottom row illustrates distinct failure cases (circled) when specific priors are removed: (a) without the Overlap Loss, roads incorrectly intersect; (b) without G1 Continuity, junctions have sharp, unnatural angles; and (c) without the Offset and Spacing Losses, curves become degenerate.}
    \label{fig:loss_comparison}
    \vspace{-2mm}
\end{figure}
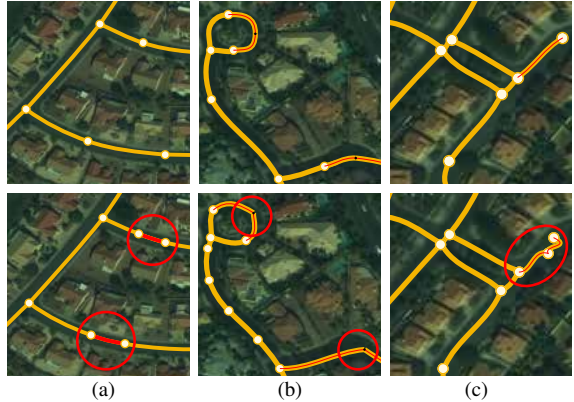

\begin{table}[h]
    \centering
    \caption{Ablation study of key components on the SpaceNet dataset. We start with the full model and ablate key components to show their impact.}
    \label{tab:ablation_tab}
    \small
    \setlength{\tabcolsep}{4pt} 
    \resizebox{\linewidth}{!}{%
    \begin{tabular}{l|cccc|c|cc}
        \toprule
        \multirow{2}{*}{Method} & \multicolumn{4}{c|}{Geometric Priors} & \multirow{2}{*}{TopoAdapt} & \multirow{2}{*}{TOPO F1$\uparrow$} & \multirow{2}{*}{APLS$\uparrow$} \\
        \cmidrule(lr){2-5}
        & $\mathcal{L}_{\text{overlap}}$ & $\mathcal{L}_{G1}$ & $\mathcal{L}_{\text{offset}}$ & $\mathcal{L}_{\text{spacing}}$ & & & \\
        \midrule
        \midrule
        Full Model & \checkmark & \checkmark & \checkmark & \checkmark & \checkmark & \textbf{84.58} & \textbf{73.48} \\
        \midrule
        w/o TopoAdapt & \checkmark & \checkmark & \checkmark & \checkmark & & 82.80 & 69.45 \\
        w/o Overlap Loss & & \checkmark & \checkmark & \checkmark & \checkmark & 83.38 & 65.95 \\
        w/o Curve Reg. & \checkmark & \checkmark & & & \checkmark & 82.57 & 65.97 \\
        w/o Geom. Priors & & & & & \checkmark & 74.58 & 51.18 \\
        Baseline($L_2$ Only) & & & & & & 70.94 & 37.10 \\
        \bottomrule
        \bottomrule
    \end{tabular}%
    }
    \vspace{-2mm}
\end{table}


\subsection{Ablation Studies}
\label{subsec:ablation_studies}

We conduct ablation studies on the SpaceNet dataset to validate our key contributions. We first establish a baseline model that only uses a basic coverage loss to optimize the B\'{e}zier graph against the segmentation mask, without any of our proposed geometric priors or the \textit{TopoAdapt} module. As shown in \cref{tab:ablation_tab}, this baseline performs poorly, confirming that a simple application of differentiable rendering is insufficient.

\textbf{Effectiveness of Geometric Priors.} Our geometric priors are crucial for plausible road geometry. Removing them entirely causes a severe degradation in performance (\cref{tab:ablation_tab}). \cref{fig:loss_comparison} visually confirms this, illustrating how ablating these priors leads to artifacts like unnatural overlaps and sharp curvatures.

\textbf{Effectiveness of Topology Editing.} The \textit{TopoAdapt} module is vital for achieving a complete and correct road network topology. Disabling this module, as shown in the corresponding ablation, causes a significant drop in metric scores as the graph can no longer be topologically refined. The baseline's poor performance further underscores that both geometric and topological enhancements are indispensable.

\section{Conclusion}
\label{sec:conclusion}
We introduced \textit{DOGE}, a new paradigm for road network extraction that reframes the task as a global optimization of a parametric B\'{e}zier Graph. By pioneering the use of differentiable rendering to align with segmentation masks, our method eliminates the need for vector GT. 
By decoupling geometric optimization (via \textit{DiffAlign}) and topological refinement (via \textit{TopoAdapt}), we reconstructs topologically accurate, geometrically smooth, and compact road networks, achieving sota results on the SpaceNet and City-Scale benchmarks.
We hope to extend the \textit{DOGE} framework to other GT-free vector reconstruction tasks.

{
    \small
    \bibliographystyle{ieeenat_fullname}
    \bibliography{main}
}


\end{document}